\documentclass{article}

\usepackage{microtype}
\usepackage{graphicx}
\usepackage{subcaption}
\usepackage{booktabs}%
\usepackage{xcolor}

\usepackage{hyperref}

\newcommand{\ours}[1]{IUFace}

\definecolor{gold}{HTML}{FAE37F}
\definecolor{silver}{HTML}{D7D7D7}
\definecolor{bronze}{HTML}{EDBA91}
\newcommand{\au}[1]{\setlength{\fboxsep}{1pt}\colorbox{gold}{#1}}
\newcommand{\ag}[1]{\setlength{\fboxsep}{1pt}\colorbox{silver}{#1}}

\usepackage[accepted]{icml2026}

\usepackage{amsmath}
\usepackage{amssymb}
\usepackage{mathtools}
\usepackage{amsthm}

\usepackage[capitalize,noabbrev]{cleveref}

\theoremstyle{plain}

\theoremstyle{definition}

\theoremstyle{remark}

\icmltitlerunning{Expressively Controllable Talking Face Generation via Identity-Aware AU Conditioning and Neutral-Reference Debiasing}

\begin{document}

\twocolumn[
  \icmltitle{Expressively Controllable Talking Face Generation via \\
    Identity-Aware AU Conditioning and Neutral-Reference Debiasing}

  \icmlsetsymbol{equal}{*}

  \begin{icmlauthorlist}
    \icmlauthor{Tri Ton}{yyy}
    \icmlauthor{Chang D. Yoo}{yyy}
  \end{icmlauthorlist}

  \icmlaffiliation{yyy}{Korea Advanced Institute of Science and Technology (KAIST)}

  \icmlcorrespondingauthor{Chang D. Yoo}{cd\_yoo@kaist.ac.kr}

  \icmlkeywords{Machine Learning, ICML}

  \vskip 0.3in
]

\printAffiliationsAndNotice{}%

\begin{abstract}
Precise control over expressive dynamics in audio-driven talking face generation remains challenging due to a lack of granular control signals. Existing methods typically rely on coarse categorical emotion labels applied across an entire video, resulting in rigid, unchanging facial poses that lack temporal diversity. To address this, we propose \ours{}, a framework that utilizes Action Units (AUs) as a continuous, multimodal conditioning signal for anatomically grounded, frame-level expression control. To ensure these movements translate naturally across diverse subjects, we introduce Identity-Aware AU Conditioning to personalize AU activations based on specific facial structures. Furthermore, our Neutral-Reference Debiasing paradigm explicitly decouples static identity from dynamic expression during training, effectively eliminating expression leakage and reference-pose tethering. Extensive evaluations demonstrate that \ours{} achieves state-of-the-art performance in fine-grained expression alignment, temporal coherence, and reliable generation.
\end{abstract}

\section{Introduction}
\label{sec:intro}

\begin{figure}[t] 
\centering
\includegraphics[width=1.\columnwidth]{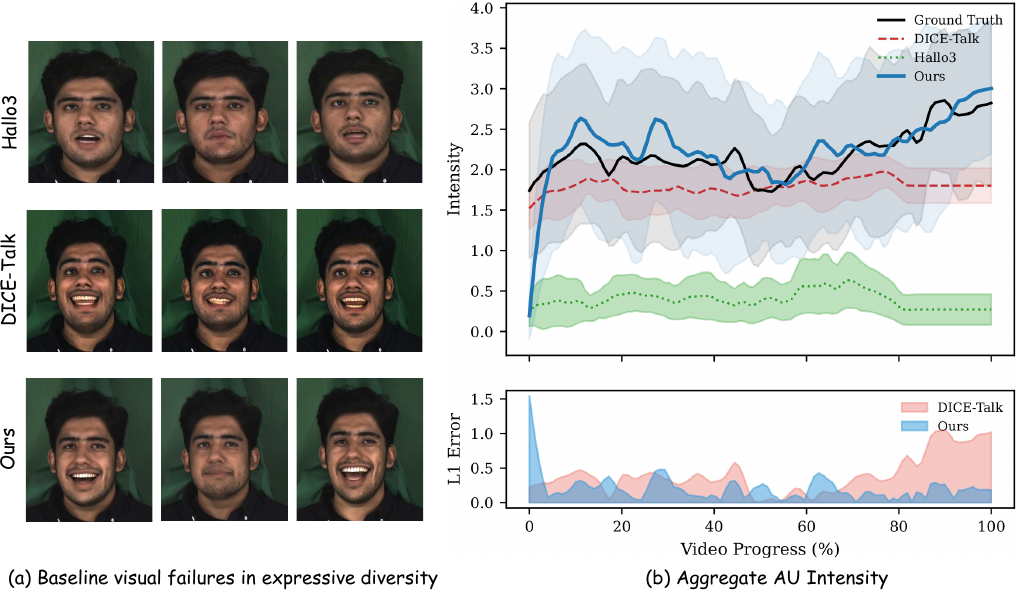} 
\caption{\textbf{Aggregate AU intensity analysis.} 
(a) Qualitative comparison of expressive diversity. (b) AU trajectories and temporal L1 error. Our method recovers natural rise-and-fall dynamics, overcoming the static bias and the absence of expression in baselines.}
\label{fig:motivation}
\vskip -0.15in
\end{figure}

Talking face generation requires the joint synthesis of phonetic synchronization and naturalistic facial affect. While diffusion-based frameworks achieve high-fidelity lip synchronization, controllable synthesis of expressive dynamics remains a bottleneck. Current architectures restrict control mechanisms to coarse emotion labels or text prompts, resulting in either a total absence of affect or a frozen, unchanging smile (Figure \ref{fig:motivation}a). Leading categorical models like DICE-Talk \cite{tan2025disentangle} exhibit a static expression bias, while non-expressive frameworks like Hallo3 \cite{cui2025hallo3} fail to reach natural expressive peaks (Figure \ref{fig:motivation}b). 

\begin{figure}[t] 
\centering
\includegraphics[width=.7\columnwidth]{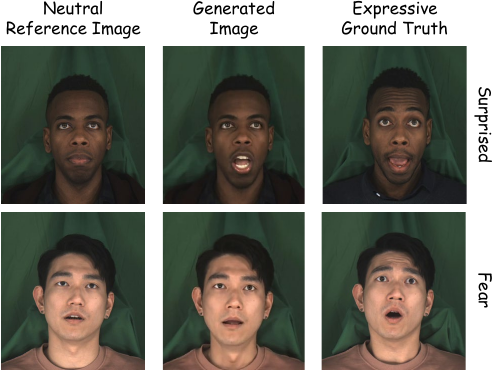} 
\caption{\textbf{Failures in Expression Disentanglement.} 
Prior methods over-rely on reference expressions. Given a neutral reference (left), baselines (middle) fail to capture the high-intensity dynamics of the ground truth (right).}
\label{fig:problem}
\vskip -0.15in
\end{figure}

To transition from rigid, short-term clips to controllable video generation, models require richer, compositional control signals. We propose utilizing Facial Action Units (AUs) as a structured, anatomically grounded multimodal control signal. Unlike text or keyframes, AUs provide a continuous temporal manifold for frame-level expression authoring, allowing creators to explicitly steer subtle micro-expressions and the natural rise-and-fall of human facial dynamics.
Realizing this control, however, is challenged by identity-expression entanglement. As shown in Figure \ref{fig:problem}, existing models over-rely on the reference identity; when provided with a neutral reference, they produce attenuated results that fail to reach the target expressive peaks.

To solve this, we introduce \ours{}, a framework for fine-grained, identity-consistent expressive control. We propose \textit{Identity-Aware AU Conditioning} to residually modulate AU activations with identity-specific features, ensuring that the same AU control signal adapts naturally to individual facial geometries. Furthermore, we introduce a \textit{Neutral-Reference Debiasing} paradigm during training to explicitly decouple static identity from dynamic motion, effectively eliminating reference-pose tethering and expression leakage.

In summary, our work makes two main contributions. First, we formulate a continuous, Identity-Aware AU control mechanism that positions Action Units as a powerful multimodal signal for personalized, frame-level expressive video directing. Second, we introduce a Neutral-Reference Debiasing paradigm that explicitly eliminates identity-expression entanglement. Together, these advancements achieve state-of-the-art temporal coherence and visual fidelity without dataset-induced reference leakage.

\section{Methodology: \ours{}}
\label{sec:method}

\begin{figure*}[t]
    \centering
    \includegraphics[width=.8\textwidth]{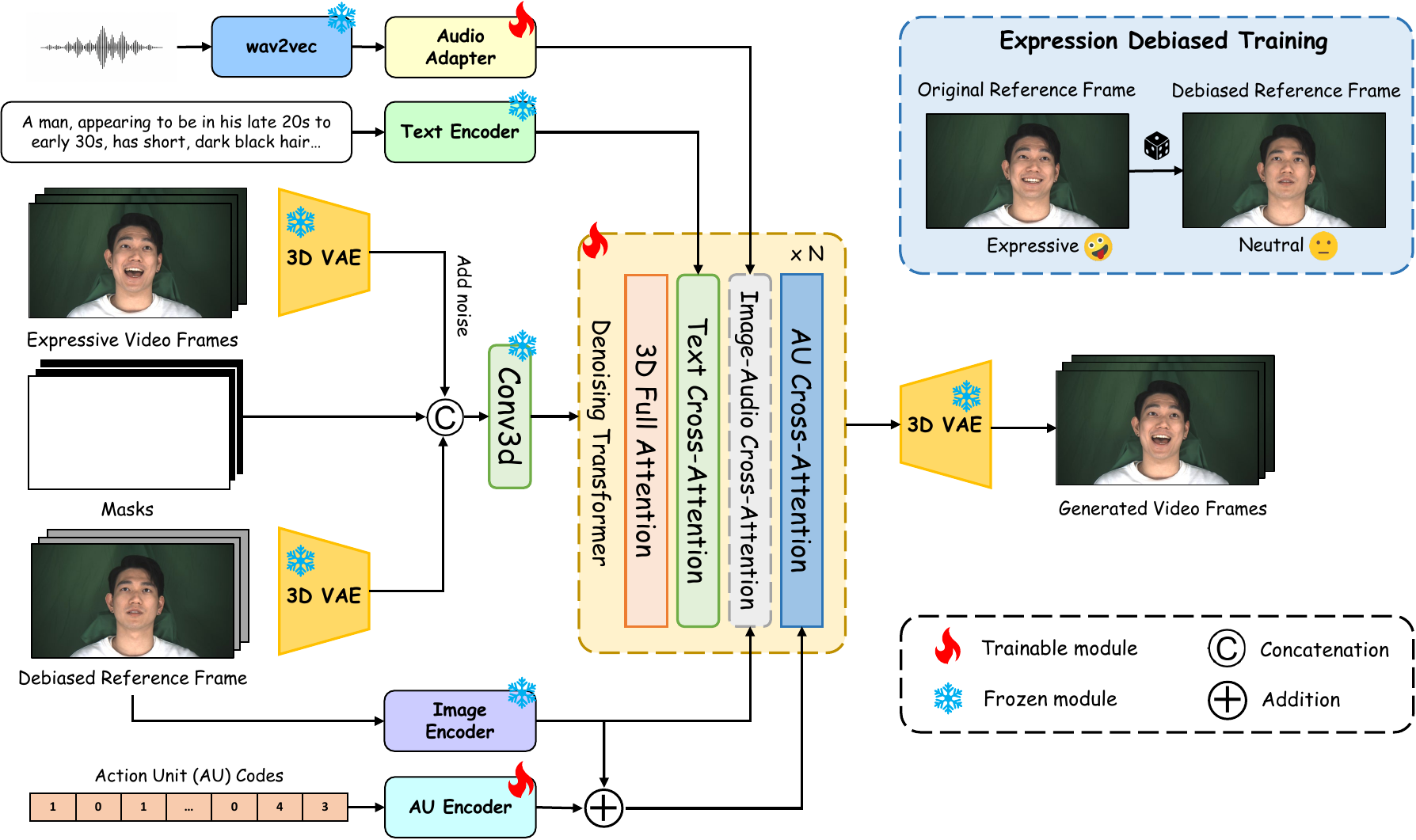}
\caption{\textbf{Overview of \ours{}.} 
\textbf{Training:} A debiased neutral reference anchors the identity, while audio and AU trajectories from an expressive video drive the dynamics. AU embeddings are fused with CLIP identity priors and injected through cross-attention.}
    \label{fig:arch}
    \vskip -0.15in
\end{figure*}

Figure \ref{fig:arch} overviews the \ours{} framework. To support reliable, long-horizon talking face generation, audio is processed via wav2vec \cite{baevski2020wav2vec} and refined through a timestep-aware vocal adapter \cite{tu2025stableavatar}, mitigating latent distribution drift over extended sequences. For fine-grained, multimodal control, we utilize continuous Action Unit (AU) trajectories. These AU features are embedded, modulated with identity-specific priors, and injected into the generative process. Concurrently, we introduce a neutral-reference debiasing strategy during training to strictly disentangle the subject's static identity from dynamic expressive biases.

\textbf{Network Architecture.}
To ensure high-fidelity motion synthesis, we build upon the Wan2.1 1.3B Video Diffusion Transformer \cite{wan2025wan}. Rather than relying on coarse text prompts, we adapt the transformer blocks to accept rich, multimodal conditioning by introducing dedicated cross-attention layers for our continuous AU signals alongside the audio embeddings. By isolating the control pathways for audio (driving phonetic sync) and AUs (driving expressive facial dynamics), the architecture provides precise, decoupled control over the final video. 
Full architectural specifications are provided in the \textit{Suppl} Sec. \ref{sec:implement}.

\noindent \textbf{Identity-Aware AU Conditioning.}
To enable fine-grained, multimodal control over facial dynamics, we utilize AU trajectories. Each AU vector $a \in \mathbb{R}^{35}$ encodes the intensity of anatomically defined muscle groups \cite{ekman1978facial}. Unlike discrete emotion labels, AUs provide a continuous, composable representation of movement. These features are embedded via a lightweight adapter: $ a' = \phi_1(a),$
where $\phi_1$ is a two-layer MLP.

While AU conditioning effectively drives motion, it is inherently identity-invariant. To ensure anatomically plausible generation, the same AU control signal must produce distinct morphological results based on the subject's unique facial geometry. We achieve this via identity-aware modulation. Rather than relying on standard face recognition embeddings optimized for discrimination, we leverage CLIP features \cite{radford2021learning}. We posit that these rich semantic priors provide a superior structural guide for generative models to reconstruct facial textures during dynamic motion. Let $f_{\text{id}} = \mathrm{MeanPool}(\mathrm{CLIP}(I_{\text{ref}})) \in \mathbb{R}^{2048}$ denote the global identity embedding, projected as:
$f'{\text{id}} = \phi_2(f{\text{id}}),$
with $\phi_2$ as a two-layer MLP. We then modulate the AU embedding residually:
{
\vskip -0.2in
\begin{equation}
c_a = a' + f'{\text{id}}.
\end{equation}
\vskip -0.1in
}
This fusion grounds the AU-driven dynamics in the specific identity (e.g., producing person-specific smile curvatures for a given AU12 intensity). The refined control signal is injected into the video latents via cross-attention:
$x_{\text{AU}} = \mathrm{Attention}(x, c_a).$
While we benchmark using video-extracted AUs, the system supports manual expression authoring via predefined 1D templates (e.g., a smile-intensity curve mapped to AU12), allowing creators to steer facial dynamics without a source video. To ensure stable early training, these attention projections are zero-initialized. Finally, the modulated AU features are integrated with the spatial and phonetic pathways:
\begin{equation}
x' = x + x_{\text{img}} + x_{\text{vocal}} + x_{\text{AU}},
\end{equation}
yielding expressive, temporally synchronized, and identity-consistent facial generation.

\noindent \textbf{Expression Debiasing via Neutral-Reference Training.}
When training on expressive datasets like MEAD \cite{wang2020mead}, models frequently entangle AU-specific facial priors with the underlying identity representation. Consequently, conditioning on an expressive reference frame causes the model to overfit, leaking residual expressions (e.g., persistent smiling) even during neutral speech. To explicitly decouple static identity from dynamic expression, we introduce a \textit{Neutral-Reference Debiasing} strategy. 

During training, rather than using an arbitrary frame from the target sequence, we randomly sample a neutral-expression frame of the same subject to serve as the identity anchor. Formally, given a neutral reference $I_{\text{ref}}^{\text{neutral}}$ and an expressive target video, the model learns the mapping:
{
\vskip -0.15in
\begin{equation}
\hat{V} = \mathcal{F}(I_{\text{ref}}^{\text{neutral}}, \mathbf{C}_{\text{AU}}, \mathbf{C}_{\text{vocal}}),
\end{equation}
\vskip -0.1in
}
where $\mathbf{C}_{\text{AU}}$ and $\mathbf{C}_{\text{vocal}}$ are the target AU and audio trajectories. This forces the network to derive all facial motion exclusively from the multimodal control signals rather than the spatial reference, effectively eliminating dataset-induced expression biases and reference-pose tethering.

\begin{table*}[t]
\centering
\scalebox{0.85}
{
\def\arraystretch{1.01}%
\aboverulesep = 0pt
\belowrulesep = 0pt
\begin{tabular}{lccccccc}
\toprule
Method & \textbf{FVD}\ $\downarrow$  & \textbf{FID}\ $\downarrow$  & \textbf{E-FID}\ $\downarrow$  & \textbf{PSNR}\ $\uparrow$ & \textbf{SSIM}\ $\uparrow$ & \textbf{F1-AU}\ $\uparrow$ & \textbf{SynC}\ $\uparrow$\\
\midrule
Hallo3 (CVPR'25) \cite{cui2025hallo3} & 277.79 & 64.53 & 4.5369 & 21.86 & 0.8292 & 0.4549 & 2.5879 \\
Sonic (CVPR'25) \cite{ji2025sonic} & \ag{192.60} & \ag{35.02} & \ag{2.2718} & 21.14 & 0.8227 & 0.4869 & 2.1125 \\
StableAvatar (arXiv) \cite{tu2025stableavatar} & 221.89 & 37.95 & 2.5441 & \ag{22.52} & \ag{0.8513} & 0.4694 & \au{2.9808} \\
DICE-Talk (ACM MM'25) \cite{tan2025disentangle} & 192.87 & 37.30 & 2.3439 & 21.65 & 0.8318 &\ag{0.5289} & 2.9063 \\
\ours{} (Ours) & \au{190.62} & \au{34.11} & \au{2.0381} & \au{24.03} & \au{0.8690} & \au{0.5895} & \ag{2.9132} \\ 
\bottomrule
\end{tabular}}
\caption{\textbf{Quantitative comparisons in the \emph{Disentanglement} Benchmark.}  We evaluate our framework on MEAD samples. We use colors to denote the \au{first}, \ag{second} respectively.}
\label{tab:main_quant}
\vspace{-3.5mm}
\end{table*}

\noindent \textbf{Classifier-Free Guidance.}
To provide creators with granular, decoupled control over emotional intensity and audio synchronization at inference time, we adopt a dual-scale classifier-free guidance (CFG) strategy. Let $\mathbf{v}_{\theta}^{\text{uncond}}$, $\mathbf{v}_{\theta}^{\text{au}}$, and $\mathbf{v}_{\theta}^{\text{cond}}$ denote the network's predictions under unconditional, AU-only, and fully conditioned (audio and AU) settings, respectively. The final extrapolated prediction is:
{
\vskip -0.15in
\begin{equation}
    \mathbf{v}_{\theta} = \mathbf{v}_{\theta}^{\text{uncond}} + w_{au} (\mathbf{v}_{\theta}^{\text{au}} - \mathbf{v}_{\theta}^{\text{uncond}}) + w_a (\mathbf{v}_{\theta}^{\text{cond}} - \mathbf{v}_{\theta}^{\text{au}}),
\end{equation}
\vskip -0.1in
}
where $w_{au}$ and $w_a$ dictate the respective guidance strengths. This formulation allows users to independently scale structural expression (via $w_{au}$) and phonetic precision (via $w_a$). Empirically, $w_{au} = 3.0$ and $w_a = 5.0$ at 50 sampling steps provide optimal anatomical coherence and sharp identity preservation without artifacts.

\section{Experiments}
\label{sec:exp}

\textbf{Experimental Setup.}
We train our model on 16,161 high-quality videos from the MEAD dataset \cite{wang2020mead}, reserving 182 videos for testing to ensure zero identity overlap. We fine-tune the Wan2.1 1.3B DiT backbone \cite{wan2025wan} on 4 NVIDIA A100 GPUs for 40,000 iterations. We evaluate generation quality using PSNR and SSIM, distributional similarity via FVD and FID, emotional fidelity via E-FID and AU-F1, and lip-sync via SyncNet. We benchmark against leading audio-driven and categorical models: Hallo3 \cite{cui2025hallo3}, Sonic \cite{ji2025sonic}, StableAvatar \cite{tu2025stableavatar}, and DICE-Talk \cite{tan2025disentangle}. Extended implementation details and cross-dataset evaluations are provided in the \textit{Suppl} Sec. \ref{sec:implement} and Sec. \ref{sec:hdtf_gen}.

\begin{figure*}[ht]
\centering
\includegraphics[width=1.7\columnwidth]{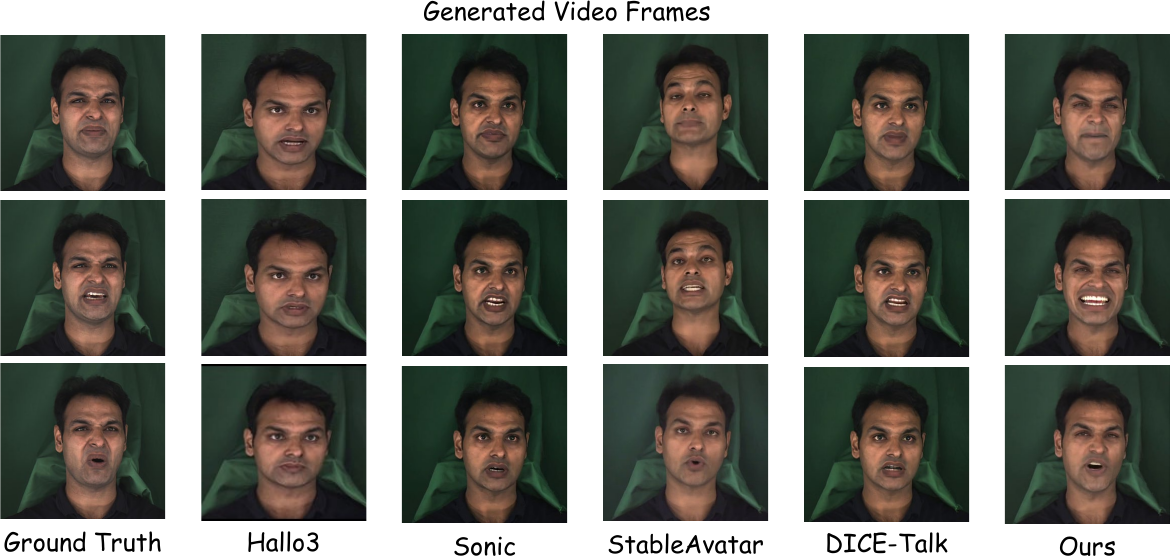} 
\caption{\textbf{Qualitative results in the \emph{Disentanglement} Benchmark.} 
A neutral reference image from a subject's neutral video is paired with audio and AU trajectories from a different expressive video.}
\label{fig:qual_1}
\vskip -0.15in
\end{figure*}

\textbf{Quantitative Results.}
To rigorously assess expressive controllability without reference-pose tethering, we use the \textbf{Disentanglement Benchmark}: a neutral reference image is paired with audio and AU trajectories from a different expressive video of the same subject. As shown in Table~\ref{tab:main_quant}, \ours{} achieves state-of-the-art performance across nearly all metrics. We obtain the lowest FVD and FID, indicating superior temporal coherence. Critically, our method leads in E-FID and AU-F1, demonstrating unmatched precision in reproducing fine-grained expression dynamics via AU control without leaking identity biases. Results on the HDTF \textit{Generalization Benchmark}, which confirm our framework's robust transferability to in-the-wild subjects and unconstrained backgrounds, are detailed in the \textit{Suppl} Sec. \ref{sec:hdtf_gen}.

\textbf{Qualitative Results.}
Visualizations in Figure \ref{fig:qual_1} compare methods in the challenging neutral-reference setting. Baselines like Hallo3 and StableAvatar exhibit blurry textures, while Sonic and DICE-Talk suffer from head jitter and asymmetric expressions. Conversely, \ours{} captures subtle micro-expressions and anatomically plausible deformations, such as symmetric cheek raising (AU6) and lip corner pulling (AU12). Our framework maintains sharp identity details and coherent temporal flow across intense expressions.

\begin{table}[t]
\centering
\scalebox{.9}{
\begin{tabular}{cccccc}
\toprule
IDA & DBE & \textbf{FVD} $\downarrow$ & \textbf{E-FID} $\downarrow$ & \textbf{F1-AU} $\uparrow$ & \textbf{SynC}\ $\uparrow$ \\
\midrule
& \checkmark   & 278.14 & 3.6821 & 0.4923 & 2.1876 \\
\checkmark  &              & 224.73 & 2.4195 & 0.5618 & 2.5876 \\
\checkmark & \checkmark & 190.62 & 2.0381 & 0.5895 & 2.9132 \\
\bottomrule
\end{tabular}
}
\caption{\textbf{Ablation study on core components on MEAD test set.} 
}
\vskip -0.2in
\label{table:ablation}
\end{table}

\begin{figure}[t] 
\centering
\includegraphics[width=.9\columnwidth]{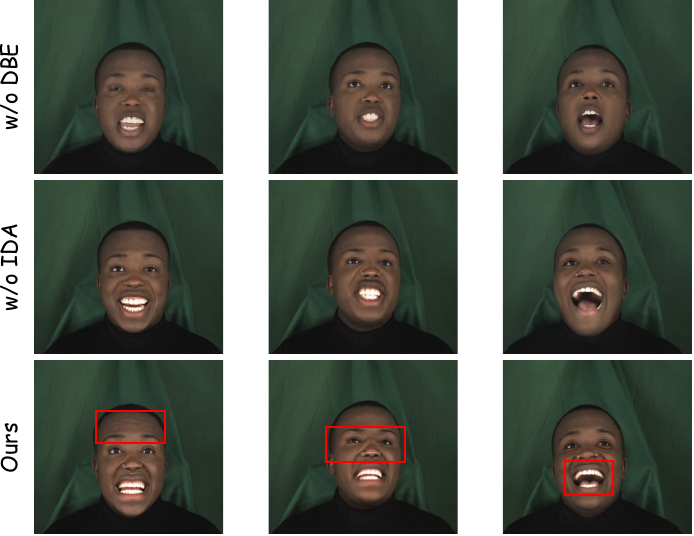} 
\caption{
\textbf{Ablation on IDA and debiasing DBE.} 
    \textbf{Middle:} w/o IDA $\rightarrow$ weak, generic expressions.
    \textbf{Top:} w/o DBE $\rightarrow$ expression leakage, identity drift. 
    \textbf{\ours{}:} sharp, personalized, debiased.}
\label{fig:ablation}
\vskip -0.15in
\end{figure}

\textbf{Ablation Studies.}
We evaluate core components in Table \ref{table:ablation} and Figure \ref{fig:ablation}. Removing Identity-Aware AU Conditioning (w/o IDA) reverts the system to an audio-driven model, causing weak, generic expressions; its inclusion significantly improves E-FID and F1-AU. Removing Emotion Debiasing (w/o DBE) means training with expressive references, which leads to expression leakage and identity drift. The full configuration achieves the best balance across all metrics, confirming that AU control and neutral-reference anchoring are essential for high-fidelity, disentangled synthesis. Further ablations on the Identity-Aware module and multi-level intensity training are provided in the \textit{Suppl} Sec. \ref{sec:detailed_ablation}.

\section{Conclusion}
\label{sec:conclusion}

We presented \ours{}, a novel framework for expressive talking face generation that achieves fine-grained control over facial dynamics while preserving strong identity consistency. By leveraging Action Units (AUs) as continuous, multimodal control signals and introducing identity-aware AU conditioning, our method enables personalized expression synthesis, overcoming the limitations of coarse categorical labels and text prompts. Furthermore, our neutral-reference debiasing strategy effectively disentangles static identity from dynamic expression, eliminating dataset-induced biases such as persistent smiling. Extensive experiments demonstrate that \ours{} achieves state-of-the-art temporal coherence, expressive fidelity, and robust generalization.

\section*{Impact Statement}

This paper presents work whose goal is to advance the field of Machine Learning. There are many potential societal consequences of our work, none of which we feel must be specifically highlighted here.

\bibliography{example_paper}

\begin{thebibliography}{31}
\providecommand{\natexlab}[1]{#1}
\providecommand{\url}[1]{\texttt{#1}}
\expandafter\ifx\csname urlstyle\endcsname\relax
  \providecommand{\doi}[1]{doi: #1}\else
  \providecommand{\doi}{doi: \begingroup \urlstyle{rm}\Url}\fi

\bibitem[Baevski et~al.(2020)Baevski, Zhou, Mohamed, and
  Auli]{baevski2020wav2vec}
Baevski, A., Zhou, Y., Mohamed, A., and Auli, M.
\newblock wav2vec 2.0: A framework for self-supervised learning of speech
  representations.
\newblock \emph{Advances in neural information processing systems}, 2020.

\bibitem[Blattmann et~al.(2023)Blattmann, Dockhorn, Kulal, Mendelevitch,
  Kilian, Lorenz, Levi, English, Voleti, Letts, et~al.]{blattmann2023stable}
Blattmann, A., Dockhorn, T., Kulal, S., Mendelevitch, D., Kilian, M., Lorenz,
  D., Levi, Y., English, Z., Voleti, V., Letts, A., et~al.
\newblock Stable video diffusion: Scaling latent video diffusion models to
  large datasets.
\newblock \emph{arXiv preprint arXiv:2311.15127}, 2023.

\bibitem[Chang et~al.(2024)Chang, Yin, Li, Tran, and
  Soleymani]{chang2024libreface}
Chang, D., Yin, Y., Li, Z., Tran, M., and Soleymani, M.
\newblock Libreface: An open-source toolkit for deep facial expression
  analysis.
\newblock In \emph{Proceedings of the IEEE/CVF winter conference on
  applications of computer vision}, pp.\  8205--8215, 2024.

\bibitem[Chen et~al.(2025)Chen, Cao, Chen, Li, and Ma]{chen2025echomimic}
Chen, Z., Cao, J., Chen, Z., Li, Y., and Ma, C.
\newblock Echomimic: Lifelike audio-driven portrait animations through editable
  landmark conditions.
\newblock In \emph{Proceedings of the AAAI Conference on Artificial
  Intelligence}, volume~39, pp.\  2403--2410, 2025.

\bibitem[Cui et~al.(2024)Cui, Li, Yao, Zhu, Shang, Cheng, Zhou, Zhu, and
  Wang]{cui2024hallo2}
Cui, J., Li, H., Yao, Y., Zhu, H., Shang, H., Cheng, K., Zhou, H., Zhu, S., and
  Wang, J.
\newblock Hallo2: Long-duration and high-resolution audio-driven portrait image
  animation.
\newblock \emph{arXiv preprint arXiv:2410.07718}, 2024.

\bibitem[Cui et~al.(2025)Cui, Li, Zhan, Shang, Cheng, Ma, Mu, Zhou, Wang, and
  Zhu]{cui2025hallo3}
Cui, J., Li, H., Zhan, Y., Shang, H., Cheng, K., Ma, Y., Mu, S., Zhou, H.,
  Wang, J., and Zhu, S.
\newblock Hallo3: Highly dynamic and realistic portrait image animation with
  video diffusion transformer.
\newblock In \emph{Proceedings of the Computer Vision and Pattern Recognition
  Conference}, 2025.

\bibitem[Ekman \& Friesen(1978)Ekman and Friesen]{ekman1978facial}
Ekman, P. and Friesen, W.~V.
\newblock Facial action coding system.
\newblock \emph{Environmental Psychology \& Nonverbal Behavior}, 1978.

\bibitem[Guo et~al.(2023)Guo, Yang, Rao, Liang, Wang, Qiao, Agrawala, Lin, and
  Dai]{guo2023animatediff}
Guo, Y., Yang, C., Rao, A., Liang, Z., Wang, Y., Qiao, Y., Agrawala, M., Lin,
  D., and Dai, B.
\newblock Animatediff: Animate your personalized text-to-image diffusion models
  without specific tuning.
\newblock \emph{arXiv preprint arXiv:2307.04725}, 2023.

\bibitem[Hong et~al.(2025)Hong, Ton, Pham, Koo, Yoon, and Yoo]{hong2025ita}
Hong, J.~W., Ton, T., Pham, T.~X., Koo, G., Yoon, S., and Yoo, C.~D.
\newblock {ITA-MDT: Image-Timestep-Adaptive Masked Diffusion Transformer
  Framework for Image-Based Virtual Try-On}.
\newblock In \emph{Proceedings of the IEEE/CVF Conference on Computer Vision
  and Pattern Recognition}, 2025.

\bibitem[Ji et~al.(2025)Ji, Hu, Xu, Zhu, Lin, He, Zhang, Luo, Chen, Lin,
  et~al.]{ji2025sonic}
Ji, X., Hu, X., Xu, Z., Zhu, J., Lin, C., He, Q., Zhang, J., Luo, D., Chen, Y.,
  Lin, Q., et~al.
\newblock Sonic: Shifting focus to global audio perception in portrait
  animation.
\newblock In \emph{Proceedings of the Computer Vision and Pattern Recognition
  Conference}, 2025.

\bibitem[Kong et~al.(2024)Kong, Tian, Zhang, Min, Dai, Zhou, Xiong, Li, Wu,
  Zhang, et~al.]{kong2024hunyuanvideo}
Kong, W., Tian, Q., Zhang, Z., Min, R., Dai, Z., Zhou, J., Xiong, J., Li, X.,
  Wu, B., Zhang, J., et~al.
\newblock Hunyuanvideo: A systematic framework for large video generative
  models.
\newblock \emph{arXiv preprint arXiv:2412.03603}, 2024.

\bibitem[Lugaresi et~al.(2019)Lugaresi, Tang, Nash, McClanahan, Uboweja, Hays,
  Zhang, Chang, Yong, Lee, et~al.]{lugaresi2019mediapipe}
Lugaresi, C., Tang, J., Nash, H., McClanahan, C., Uboweja, E., Hays, M., Zhang,
  F., Chang, C.-L., Yong, M.~G., Lee, J., et~al.
\newblock Mediapipe: A framework for building perception pipelines.
\newblock \emph{arXiv preprint arXiv:1906.08172}, 2019.

\bibitem[Luu et~al.(2026)Luu, Ton, Nguyen, Nguyen, Binh, Nguyen, Duong,
  Sonntag, Nguyen, Ngo, and Yoo]{luu2026progress}
Luu, T.~M., Ton, T., Nguyen, D.~M., Nguyen, X.~T., Binh, N.~G., Nguyen, Q.~T.,
  Duong, V., Sonntag, D., Nguyen, D. M.~H., Ngo, V.~A., and Yoo, C.~D.
\newblock {Progress-Guided Preference Generation for Aligning Generative
  Control Policies}.
\newblock In \emph{Conference on Robot Learning}, 2026.

\bibitem[Nguyen et~al.(2026)Nguyen, Ton, Choe, Luu, and Yoo]{nguyen2026fast}
Nguyen, T., Ton, T., Choe, H., Luu, T.~M., and Yoo, C.~D.
\newblock {Fast and Highly Expressive Policy Learning for Offline Reinforcement
  Learning via Bootstrapped Flow Q-Learning}.
\newblock In \emph{International Conference on Machine Learning}, 2026.

\bibitem[Peebles \& Xie(2023)Peebles and Xie]{peebles2023scalable}
Peebles, W. and Xie, S.
\newblock Scalable diffusion models with transformers.
\newblock \emph{ICCV}, 2023.

\bibitem[Pham et~al.(2025)Pham, Ton, and Yoo]{pham2024mdsgen}
Pham, T.~X., Ton, T., and Yoo, C.~D.
\newblock {MDSGen: Fast and Efficient Masked Diffusion Temporal-Aware
  Transformers for Open-Domain Sound Generation}.
\newblock In \emph{International Conference on Learning Representations}, 2025.

\bibitem[Radford et~al.(2021)Radford, Kim, Hallacy, Ramesh, Goh, Agarwal,
  Sastry, Askell, Mishkin, Clark, et~al.]{radford2021learning}
Radford, A., Kim, J.~W., Hallacy, C., Ramesh, A., Goh, G., Agarwal, S., Sastry,
  G., Askell, A., Mishkin, P., Clark, J., et~al.
\newblock Learning transferable visual models from natural language
  supervision.
\newblock In \emph{International conference on machine learning}, 2021.

\bibitem[Singer et~al.(2022)Singer, Polyak, Hayes, Yin, An, Zhang, Hu, Yang,
  Ashual, Gafni, et~al.]{singer2022make}
Singer, U., Polyak, A., Hayes, T., Yin, X., An, J., Zhang, S., Hu, Q., Yang,
  H., Ashual, O., Gafni, O., et~al.
\newblock Make-a-video: Text-to-video generation without text-video data.
\newblock \emph{arXiv preprint arXiv:2209.14792}, 2022.

\bibitem[Tan et~al.(2025)Tan, Lin, Xu, Xu, Hu, Ji, Zhu, Wang, and
  Fu]{tan2025disentangle}
Tan, W., Lin, C., Xu, C., Xu, F., Hu, X., Ji, X., Zhu, J., Wang, C., and Fu, Y.
\newblock Disentangle identity, cooperate emotion: Correlation-aware emotional
  talking portrait generation.
\newblock \emph{arXiv preprint arXiv:2504.18087}, 2025.

\bibitem[Team et~al.(2023)Team, Anil, Borgeaud, Alayrac, Yu, Soricut,
  Schalkwyk, Dai, Hauth, Millican, et~al.]{team2023gemini}
Team, G., Anil, R., Borgeaud, S., Alayrac, J.-B., Yu, J., Soricut, R.,
  Schalkwyk, J., Dai, A.~M., Hauth, A., Millican, K., et~al.
\newblock Gemini: a family of highly capable multimodal models.
\newblock \emph{arXiv preprint arXiv:2312.11805}, 2023.

\bibitem[Tee et~al.(2026)Tee, Yoon, Yoon, Ton, and Yoo]{tee2026diffusion}
Tee, J. T.~J., Yoon, H.~S., Yoon, S., Ton, T., and Yoo, C.~D.
\newblock {Diffusion Negative Preference Optimization Made Simple}.
\newblock In \emph{International Conference on Learning Representations}, 2026.

\bibitem[Ton et~al.(2024)Ton, Hong, Eom, Shim, Kim, and Yoo]{ton2024dual}
Ton, T., Hong, J.~W., Eom, S., Shim, J.~Y., Kim, J., and Yoo, C.~D.
\newblock {Zero-Shot Dual-Path Integration Framework for Open-Vocabulary 3D
  Instance Segmentation}.
\newblock In \emph{Proceedings of the IEEE/CVF Conference on Computer Vision
  and Pattern Recognition Workshops}, 2024.

\bibitem[Ton et~al.(2025)Ton, Hong, and Yoo]{ton2025taro}
Ton, T., Hong, J.~W., and Yoo, C.~D.
\newblock {TARO: Timestep-Adaptive Representation Alignment with Onset-Aware
  Conditioning for Synchronized Video-to-Audio Synthesis}.
\newblock In \emph{Proceedings of the IEEE/CVF International Conference on
  Computer Vision}, 2025.

\bibitem[Ton et~al.(2026)Ton, Lim, Tee, and Yoo]{ton2026focusflow}
Ton, T., Lim, S., Tee, J. T.~J., and Yoo, C.~D.
\newblock {Training-Free 3D Editing via Feature-Divergence Localization and
  Trajectory Correction}.
\newblock In \emph{Advances in Neural Information Processing Systems}, 2026.

\bibitem[Tu et~al.(2025)Tu, Pan, Huang, Han, Xing, Dai, Luo, Wu, and
  Jiang]{tu2025stableavatar}
Tu, S., Pan, Y., Huang, Y., Han, X., Xing, Z., Dai, Q., Luo, C., Wu, Z., and
  Jiang, Y.-G.
\newblock Stableavatar: Infinite-length audio-driven avatar video generation.
\newblock \emph{arXiv preprint arXiv:2508.08248}, 2025.

\bibitem[Wan et~al.(2025)Wan, Wang, Ai, Wen, Mao, Xie, Chen, Yu, Zhao, Yang,
  et~al.]{wan2025wan}
Wan, T., Wang, A., Ai, B., Wen, B., Mao, C., Xie, C.-W., Chen, D., Yu, F.,
  Zhao, H., Yang, J., et~al.
\newblock Wan: Open and advanced large-scale video generative models.
\newblock \emph{arXiv preprint arXiv:2503.20314}, 2025.

\bibitem[Wang et~al.(2020)Wang, Wu, Song, Yang, Wu, Qian, He, Qiao, and
  Loy]{wang2020mead}
Wang, K., Wu, Q., Song, L., Yang, Z., Wu, W., Qian, C., He, R., Qiao, Y., and
  Loy, C.~C.
\newblock Mead: A large-scale audio-visual dataset for emotional talking-face
  generation.
\newblock In \emph{European conference on computer vision}, 2020.

\bibitem[Wei et~al.(2024)Wei, Yang, and Wang]{wei2024aniportrait}
Wei, H., Yang, Z., and Wang, Z.
\newblock Aniportrait: Audio-driven synthesis of photorealistic portrait
  animation.
\newblock \emph{arXiv preprint arXiv:2403.17694}, 2024.

\bibitem[Xu et~al.(2024)Xu, Li, Su, Shang, Zhang, Liu, Wang, Yao, and
  Zhu]{xu2024hallo}
Xu, M., Li, H., Su, Q., Shang, H., Zhang, L., Liu, C., Wang, J., Yao, Y., and
  Zhu, S.
\newblock Hallo: Hierarchical audio-driven visual synthesis for portrait image
  animation.
\newblock \emph{arXiv preprint arXiv:2406.08801}, 2024.

\bibitem[Yang et~al.(2024)Yang, Teng, Zheng, Ding, Huang, Xu, Yang, Hong,
  Zhang, Feng, et~al.]{yang2024cogvideox}
Yang, Z., Teng, J., Zheng, W., Ding, M., Huang, S., Xu, J., Yang, Y., Hong, W.,
  Zhang, X., Feng, G., et~al.
\newblock Cogvideox: Text-to-video diffusion models with an expert transformer.
\newblock \emph{arXiv preprint arXiv:2408.06072}, 2024.

\bibitem[Zhang et~al.(2021)Zhang, Li, Ding, and Fan]{zhang2021flow}
Zhang, Z., Li, L., Ding, Y., and Fan, C.
\newblock Flow-guided one-shot talking face generation with a high-resolution
  audio-visual dataset.
\newblock In \emph{Proceedings of the IEEE/CVF conference on computer vision
  and pattern recognition}, 2021.

\end{thebibliography}
\bibliographystyle{icml2026}

\newpage
\appendix
\onecolumn
This supplementary material provides comprehensive technical details, an extended literature review, and expanded experimental results to support the main paper. Specifically, Section \ref{sec:related} provides a broader discussion of related works. Section \ref{sec:implement} outlines our network architecture and implementation details, while Section \ref{sec:runtime} offers a comparative analysis of model parameters and computational efficiency. Further empirical evidence is detailed in subsequent sections: Section \ref{sec:hdtf_gen} presents cross-dataset generalization results on the HDTF dataset, Section \ref{sec:detailed_ablation} provides additional granular ablation studies, and Section \ref{sec:more_quan} includes extended standard-setting evaluations on the MEAD dataset. 
Finally, we discuss current framework limitations in Section \ref{sec:limitations}, outline future directions in Section \ref{sec:future}, and address ethical considerations regarding the responsible deployment of expressive generative models in Section \ref{sec:ethical}.

\section{Related Works}
\label{sec:related}

\noindent \textbf{Diffusion-based Video Generation.}
Diffusion models have transformed video synthesis by scaling image priors to temporal domains. Early works like Make-A-Video \cite{singer2022make} and AnimateDiff \cite{guo2023animatediff} adapt diffusion via spatial-temporal factorization and motion modules, enabling zero-shot animation from unlabeled videos. Latent diffusion further improves efficiency. Stable Video Diffusion \cite{blattmann2023stable} uses staged pretraining with curated data to achieve state-of-the-art text-to-video and image-to-video quality. Recent large-scale Diffusion Transformers (DiTs) \cite{peebles2023scalable, yang2024cogvideox, wan2025wan, kong2024hunyuanvideo} push boundaries with billion-parameter models and 3D causal VAEs to enable high-fidelity, controllable generation. These advancements in scalable motion modeling directly inspire modern talking head synthesis, where DiT backbones are conditioned on audio and fine-grained signals to produce expressive, temporally coherent facial dynamics.

\noindent \textbf{Talking Face Generation.}
Audio-driven talking face generation has progressed from fine-tuning image-based diffusion models with temporal modules to leveraging pretrained video backbones for enhanced realism and scalability. \cite{wei2024aniportrait, xu2024hallo, cui2024hallo2, chen2025echomimic} adapt AnimateDiff \cite{guo2023animatediff} by inserting temporal modules and editable landmark conditioning, enabling synchronized lip and expression animation without intermediate representations. More recent methods exploit large video diffusion transformers: Hallo3 \cite{cui2025hallo3} introduces identity reference networks and causal 3D VAEs for dynamic, non-frontal portrait animation. Sonic \cite{ji2025sonic} enhances global audio perception via intra/inter-clip disentanglement and context-aware learning. StableAvatar \cite{tu2025stableavatar} enables infinite-length generation through timestep-aware audio adapters and sliding-window fusion to prevent latent drift. To boost expressiveness, \cite{cui2024hallo2, tan2025disentangle} incorporate prompts and emotion labels, but suffer from coarse control and identity leakage. In contrast, our work achieves fine-grained, identity-aware emotional controllability via anatomically precise AU conditioning and neutral-reference debiasing.

\section{Implementation details}
\label{sec:implement}

The Denoising Transformer Network consists of a 30-layer DiT backbone adapted from Wan2.1-Fun-V1.1-1.3B \cite{wan2025wan}. 
We increase the input channel dimension to 36 to accommodate the concatenation of masked latents and binary mask channels. 
The transformer uses hidden dimension 1536, 12 attention heads, intermediate FFN size 8960, and 3D patching of size $1{\times}2{\times}2$. 
Training is performed with Flow-Matching Euler scheduler, mixed-precision bf16, effective batch size 4 (1 per GPU) on 4$\times$A100-80GB, learning rate $2{\times}10^{-5}$, and AdamW optimizer.

During training, we employ a reconstruction loss in latent space. 
To better supervise the most challenging facial regions, we derive two binary masks from MediaPipe \cite{lugaresi2019mediapipe}: a precise inner-lip mask and a broader full-face mask. 
At each training iteration, with probability 40\% we apply no regional weighting, with 10\% probability we weight only the full-face region, and with 50\% probability we weight only the lip region by element-wise multiplying the per-pixel squared error with the corresponding mask before global averaging. 
This stochastic region-focused supervision substantially improves lip synchronization accuracy and upper-face expressiveness.

Text prompts, generated by Gemini \cite{team2023gemini}, are strictly limited to describing static appearance (e.g., hair color, lighting) to ensure that expressive dynamics are driven exclusively by AU and audio conditioning without confounding linguistic cues.

\begin{table}[ht]
\centering
\resizebox{.7\hsize}{!}{
\begin{tabular}{lcccc}
\toprule
Method & \textbf{Params (B)} & \textbf{GPU Mem (GB)} & \textbf{Speed (s)} & \textbf{F1-AU}$\uparrow$ \\
\midrule
Hallo3 \cite{cui2025hallo3} & 14.50  & 49.77 & 967.00 & 0.4549  \\
Sonic \cite{ji2025sonic} & 1.58  & 19.99 & 118.26 & 0.4869 \\
StableAvatar \cite{tu2025stableavatar} & 1.75 & 30.53 & 578.10 & 0.4694 \\
DICE-Talk \cite{tu2025stableavatar} & 1.63 & 22.18 & 176.94 & 0.5289 \\
\ours{} (Ours) & 1.89 & 31.44 & 816.59 & 0.5895 \\
\bottomrule
\end{tabular}}
\vskip 0.1in
\caption{Model size and inference efficiency on an A100-80GB GPU (81 frames).}
\vskip -0.15in
\label{table:inference}
\end{table}

\section{Parameters and Runtime Comparison}
\label{sec:runtime}

Among methods with fewer than 2B parameters, \ours{} achieves the highest F1-AU score by a clear margin  while keeping both parameter count and memory footprint highly competitive. 
The increased inference time primarily stems from the deliberate design choice of using full-attention 3D transformer blocks with explicit AU cross-attention and identity-aware conditioning, which are essential for the fine-grained emotional control and identity preservation demonstrated throughout our results in the main paper and supplementary videos. 
This represents a favorable trade-off between expressive quality and efficiency for applications that prioritize realism and controllability over performance.

\section{HDTF Generalization Results}
\label{sec:hdtf_gen}

To evaluate the generalizability of our framework to in-the-wild subjects and unconstrained backgrounds, this section presents our cross-dataset evaluation on the HDTF dataset \cite{zhang2021flow}.

\begin{table*}[ht]
\centering
\scalebox{0.85}{
\begin{tabular}{lccccccc}
\toprule
Method & \textbf{FVD} $\downarrow$ & \textbf{FID} $\downarrow$ & \textbf{E-FID} $\downarrow$ & \textbf{PSNR} $\uparrow$ & \textbf{SSIM} $\uparrow$ & \textbf{F1-AU} $\uparrow$ & \textbf{SynC} $\uparrow$\\
\midrule
Hallo3 (CVPR'25) \cite{cui2025hallo3} & 153.39 & 13.96 & 0.2789 & 20.85 & 0.7735 & 0.6342 & 3.9839 \\
Sonic (CVPR'25) \cite{ji2025sonic} & 163.18 & 12.16 & 0.2981 & 21.14 & 0.7795 & 0.6385 & \ag{5.0812} \\
StableAvatar (arXiv) \cite{tu2025stableavatar} & 167.40 & \ag{10.59} & \ag{0.2378} & 20.11 & 0.7727 & 0.5788 & 5.0727 \\
DICE-Talk (ACM MM'25) \cite{tan2025disentangle} & 178.50 & 12.15 & 0.2415 & \ag{21.28} & \au{0.7848} & \ag{0.6420} & \au{5.3501} \\
\ours{} (Ours) & \au{147.10} & \au{10.40} & \au{0.2254} & \au{21.73} & \ag{0.7841} & \au{0.6500} & 5.1650 \\
\bottomrule
\end{tabular}}
\caption{\textbf{Quantitative comparisons in the \emph{Generalization} Benchmark.} We evaluate our framework on the full HDTF dataset.}
\label{tab:main_quant_hdtf}
\end{table*}

\begin{figure*}[h]
\centering
\includegraphics[width=.8\columnwidth]{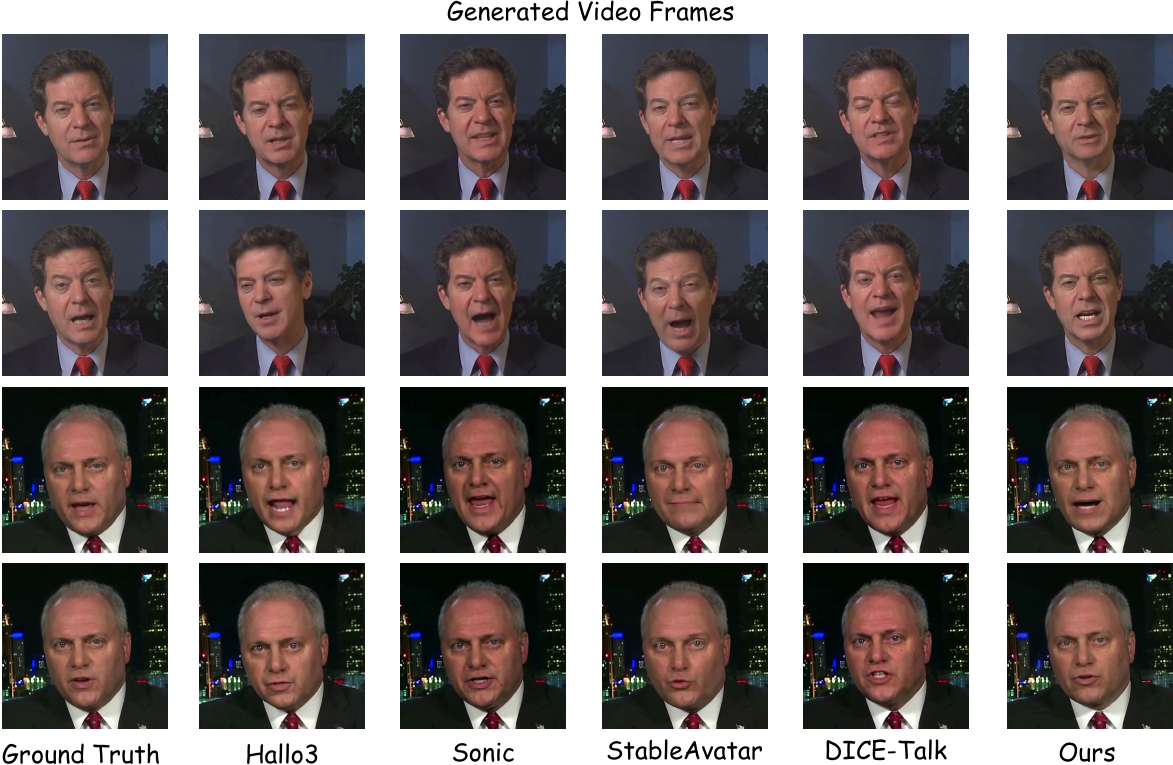}
\caption{\textbf{Qualitative results in the \emph{Generalization} Benchmark.}
A reference image is paired with audio and AU trajectories from the same expressive video.}
\label{fig:qual_hdtf}
\vskip -0.15in
\end{figure*}

\noindent \textbf{Quantitative Results.}
Quantitative results on HDTF in Table \ref{tab:main_quant_hdtf} demonstrate the robust transferability of our framework to in-the-wild identities and unconstrained backgrounds. We achieve the best temporal coherence and visual quality with the lowest FVD and FID scores. Notably, our method sets a new state of the art in E-FID and F1-AU, indicating that identity-aware AU conditioning effectively preserves expression fidelity during motion. While DICE-Talk \cite{tan2025disentangle} and Sonic \cite{ji2025sonic} maintain slight leads in SSIM and SynC, our framework achieves the highest PSNR and competitive performance across all auxiliary metrics. These results confirm that leveraging CLIP-based semantic structural priors successfully resolves the trade-off between phonetic synchronization and anatomical consistency. These metrics remain stable when evaluated on the full HDTF dataset, confirming the statistical reliability of our generalization claims.

\noindent \textbf{Qualitative Results.}
Figure \ref{fig:qual_hdtf} illustrates qualitative comparisons on the HDTF dataset, where the reference image and driving signals are sampled from the same expressive sequence. While existing methods often struggle with unconstrained backgrounds, \ours{} maintains superior structural integrity and temporal coherence. Specifically, Hallo3 \cite{cui2025hallo3} and StableAvatar \cite{tu2025stableavatar} exhibit noticeable blurring in the perioral region and tend to produce rigid facial dynamics. Although Sonic \cite{ji2025sonic} and DICE-Talk \cite{tan2025disentangle} capture motion more effectively, they occasionally suffer from asymmetric mouth deformations and unnatural head orientations during high-intensity speech. In contrast, our framework employs identity-aware AU conditioning to preserve subject-specific traits, such as unique smile curvatures and eye geometry, while mitigating identity drift.

\section{Detailed Ablation Studies}
\label{sec:detailed_ablation}

This section extends the core ablation study presented in the main paper by providing two additional analyses on our finer-grained design choices. 

\begin{table}[t]
\centering
\scalebox{0.9}{
\begin{tabular}{lcccc}
\toprule
        & \textbf{FVD} $\downarrow$ & \textbf{E-FID} $\downarrow$ & \textbf{F1-AU} $\uparrow$ & \textbf{SynC}\ $\uparrow$ \\
\midrule
w/o IA   & 267.38 & 3.5914 & 0.5012 & 2.3127 \\
w IA     & 190.62 & 2.0381 & 0.5895 & 2.9132      \\
\midrule
Level 3 Only  & 231.19 & 2.8472 & 0.5421 & 2.6014 \\
All Levels       & 190.62 & 2.0381 & 0.5895 & 2.9132    \\
\bottomrule
\end{tabular}
}
\caption{\textbf{Quantitative results of ablation study on detailed modules on MEAD test set.} 
}
\vskip -0.25in
\label{table:ablation_fine}
\end{table}

\subsection{Identity-Aware Modulation}
Table \ref{table:ablation_fine} presents the impact of the Identity-Aware (IA) module. Without IA, the model utilizes AU conditioning but lacks subject-specific modulation, effectively removing the identity features from the expression control logic. This configuration leads to identity drift and visibly degraded facial dynamics. Introducing IA markedly improves identity preservation and expression fidelity, demonstrating that modeling subject-specific facial priors is essential for high-quality talking head synthesis.

\subsection{Training Intensity Diversity}
The MEAD dataset categorizes emotion intensities from level 1 (mild) to level 3 (strong). Training exclusively on extreme expressions (Level 3 Only) restricts the model to high muscle activations, causing identity entanglement and poor subtle dynamics. Transitioning to All Levels enables a continuous mapping of AU intensities, which significantly improves E-FID and F1-AU as shown in Table \ref{table:ablation_fine}. This multi-level strategy allows the model to decouple static identity from emotional variance and capture subtle facial movements.

\subsection{Independent AU Evaluation}

To verify that the observed performance gains are not artifacts of utilizing OpenFace for both training and evaluation, we assess the generated expressions using an independent AU estimator.
\begin{table}[h]
\centering
\begin{tabular}{lcc}
\toprule
Method & LibreFace F1-AU & OpenFace F1-AU \\
\midrule
Sonic (CVPR'25) & 0.4512 & 0.4869 \\
DICE-Talk (ACM MM'25) & 0.4934 & 0.5289 \\
\ours{} (Ours) & 0.5608 & 0.5895 \\
\bottomrule
\end{tabular}
\caption{LibreFace vs. OpenFace AU-F1 on MEAD test set. \ours{} maintains a decisive margin under an independent estimator.}
\vskip -0.25in
\label{tab:independent_au}
\end{table}
As shown in Table \ref{tab:independent_au}, \ours{} maintains a decisive margin over all baselines under LibreFace~\cite{chang2024libreface}, confirming true anatomical accuracy independent of OpenFace's algorithmic biases.

\subsection{Identity Encoder Ablation}

To evaluate our design choices, we compare the use of CLIP versus ArcFace as the identity embedding within the Identity-Aware AU Conditioning module. 

\begin{table}[h]
\centering
\begin{tabular}{lccc}
\toprule
Identity Embedding & FVD $\downarrow$ & PSNR $\uparrow$ & SSIM $\uparrow$ \\
\midrule
ArcFace (ResNet-50) & 215.42 & 22.80 & 0.831 \\
CLIP ViT-L/14 (Ours) & 190.62 & 24.03 & 0.869 \\
\bottomrule
\end{tabular}
\caption{CLIP vs. ArcFace identity embeddings on MEAD Disentanglement Benchmark.}
\label{tab:id_encoder_ablation}
\vskip -0.15in
\end{table}

CLIP significantly outperforms ArcFace because high-fidelity generative reconstruction requires rich semantic texture priors rather than purely discriminative features. Because ArcFace is trained specifically to discard these intra-class variations, relying on it leads to flattened, less detailed facial textures during generation.

\subsection{Identity Preservation Under Debiasing}

We measure ArcFace cosine similarity between generated frames and the identity reference. Under the neutral-reference setting, similarity is 0.82; under the expressive-reference setting, it is 0.81. This minimal 0.01 variance confirms that Neutral-Reference Debiasing successfully locks identity representation independent of the driving expression intensity.

\section{More MEAD Results}
\label{sec:more_quan}

\begin{table*}[t]
\centering
\scalebox{0.9}
{\begin{tabular}{lcccccccc}
\toprule
Method & \textbf{FVD}\ $\downarrow$  & \textbf{FID}\ $\downarrow$  & \textbf{E-FID}\ $\downarrow$  & \textbf{PSNR}\ $\uparrow$ & \textbf{SSIM}\ $\uparrow$ &\textbf{LPIPS}\ $\downarrow$ & \textbf{F1-AU}\ $\uparrow$ & \textbf{SynC}\ $\uparrow$\\
\midrule
Hallo3 (CVPR'25) \cite{cui2025hallo3} & 137.57 & 51.01 & 1.9575 & 24.23 & 0.8533 & 0.2529 & 0.6627 & \ag{3.1806} \\
Sonic (CVPR'25) \cite{ji2025sonic} & \ag{109.71} & \au{18.72} & \ag{1.2161} & 24.07 & 0.8912 & \au{0.1055} & \ag{0.6703}  & 2.7603 \\
StableAvatar (arXiv) \cite{tu2025stableavatar} & 156.90 & 22.78 & 1.4898 & \ag{26.12} & 0.8946 & 0.1455 & 0.6262 & 3.1075 \\
DICE-Talk (ACM MM'25) \cite{tan2025disentangle} & 139.27 & 24.12 & 1.8593 & 25.12 & \ag{0.8968} & \ag{0.1105} & 0.6680 & \au{3.2597} \\
\ours{} (Ours) & \au{97.73} & \ag{22.40} & \au{1.0927} & \au{28.79} & \au{0.9210} & 0.1205 & \au{0.6918} & 3.1092 \\ 
\bottomrule
\end{tabular}}
\caption{\textbf{Quantitative comparisons in the second setting.}  We evaluate our framework on MEAD samples.}
\label{tab:main_quant_nodebiased}
\end{table*}

\begin{figure*}[h]
\centering
\includegraphics[width=.7\columnwidth]{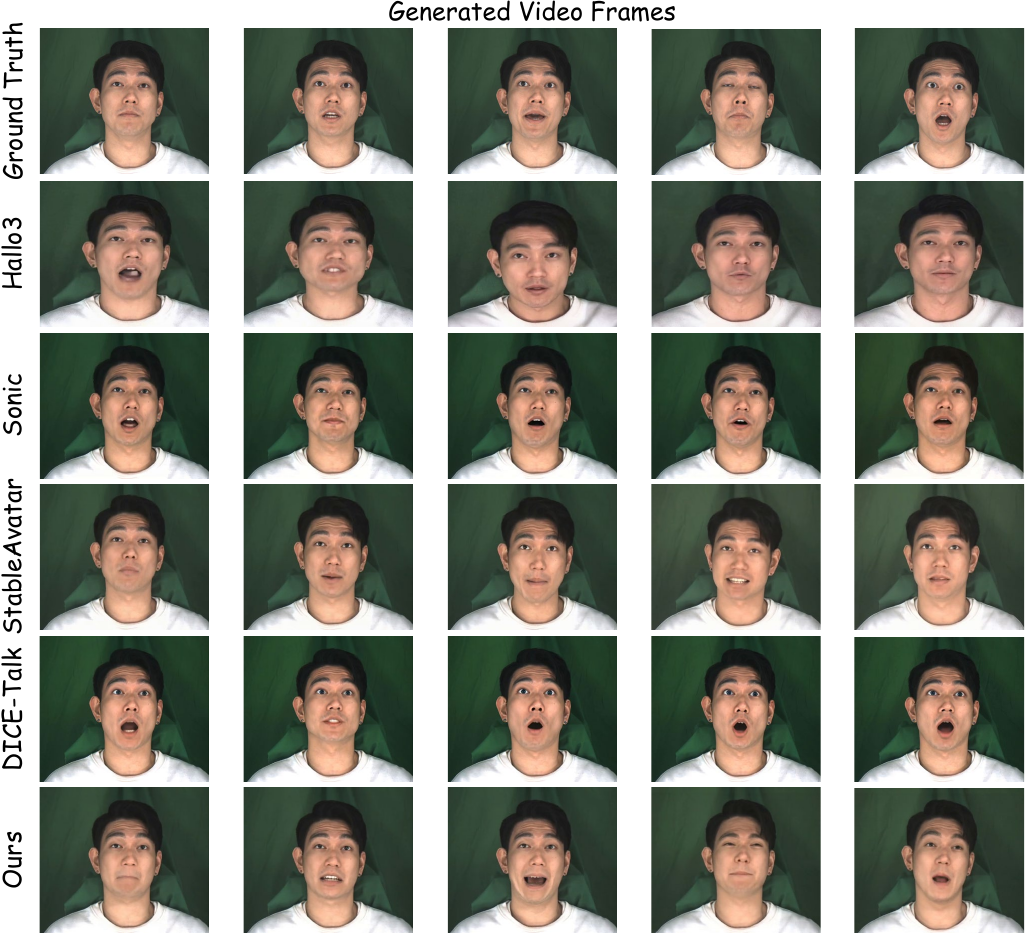} 
\caption{\textbf{Qualitative results in the MEAD dataset} 
A neutral reference image is paired with audio and AU trajectories from the same expressive video. 
}
\label{fig:qual_hdtf_append}
\vskip -0.25in
\end{figure*}

As shown in Table \ref{tab:main_quant_nodebiased}, our framework achieves state-of-the-art performance across all metrics on MEAD under the standard setting, where the first frame of the target video serves as the reference image. 
We obtain the lowest FVD and E-FID, confirming superior temporal dynamics and expression fidelity. 
Our method also leads in PSNR, SSIM, and AU-F1, demonstrating unmatched structural quality and fine-grained expression control. 
While FID is competitive, the overall gains, particularly in LPIPS and Sync, validate the robustness of our identity-aware AU conditioning even when expressive references are available. 
These results confirm general applicability across both debiased and standard training paradigms.

Figure \ref{fig:qual_hdtf_append} presents qualitative comparisons under the standard setting, where the first frame of the target video serves as the reference image. 
Our method continues to outperform baselines, generating sharper facial details, more natural mouth deformations, and smoother expression transitions even when the reference contains expressive cues. 
Hallo3 \cite{cui2025hallo3}, and StableAvatar \cite{tu2025stableavatar} produce overly rigid expressions and blurred lip regions, while Sonic \cite{ji2025sonic} and DICE-Talk \cite{tan2025disentangle} frequently exhibit asymmetric mouth deformations and unnatural head jitter when attempting to transition from the initial expressive pose. These comparisons confirm that \ours{} consistently outperforms all evaluated baselines in maintaining high-fidelity, anatomically coherent facial dynamics under the standard setting.

\section{Limitations}
\label{sec:limitations}

Our framework faces challenges in generating multiple humans in reference images, as the identity-aware module currently processes one subject at a time. Extending this to multi-person scenes requires robust identity disentanglement and spatial attention mechanisms, which we leave for future exploration.
The use of the diffusion model results in lower operational efficiency, with slower inference times due to the computational complexity of the denoising process.
We plan to investigate latent consistency models, distillation into flow-based generators to achieve sub-second inference while preserving fidelity.

\section{Future Work}
\label{sec:future}

Several directions remain for extending \ours{}. To reduce the inference cost noted in Section \ref{sec:limitations}, bootstrapped single-step flow learning, developed for offline reinforcement learning \cite{nguyen2026fast}, could be adapted to the flow-matching Wan2.1 backbone \cite{wan2025wan}. For multi-person scenes, zero-shot integration of 2D and 3D open-vocabulary mask proposals \cite{ton2024dual} could localize each subject, routing AU and audio conditions to the corresponding speaker. The mean-pooled CLIP identity embedding could also be replaced by timestep-adaptive aggregation of image-encoder features \cite{hong2025ita} to better retain person-specific details under intense expressions. Furthermore, Feature-Divergence Localization \cite{ton2026focusflow}, which gates the velocity field with intrinsic cross-attention, could enable training-free expression editing of existing videos confined to the edited facial regions, while its phase-scheduled integration motivates timestep-dependent guidance scales in place of our fixed $w_{au}$ and $w_a$. Artifacts such as head jitter and expression leakage could be suppressed through single-network negative preference optimization \cite{tee2026diffusion}, with preference labels derived from real recordings without human annotation via progress-guided preference generation \cite{luu2026progress}. Finally, onset-aware video-to-audio synthesis \cite{ton2025taro} and efficient masked diffusion transformers for sound generation \cite{pham2024mdsgen} could produce non-verbal vocalizations, such as laughter, synchronized with the generated AU dynamics, moving toward jointly generated audio-visual talking faces.

\section{Ethical Concern}
\label{sec:ethical}

Our method synthesizes highly realistic talking-face videos from a single reference image using arbitrary driving audio and fine-grained Action Unit trajectories. This capability supports valuable applications, including virtual avatars, automated dubbing, film post-production, and assistive technologies for speech-impaired individuals. At the same time, it introduces the risk of misuse for creating non-consensual videos of real people, impersonation, or spreading misinformation through synthetic media.
To minimize potential societal harm, we strongly recommend that any practical deployment incorporate robust deepfake detection models specifically trained to recognize signatures of AU-conditioned synthesis, as well as imperceptible authentication watermarks embedded in the generated videos.

\end{document}